\documentclass[11pt]{article}

\usepackage[preprint]{acl}
\makeatletter
\AtBeginDocument{\@ifundefined{runningpagewiselinenumbers}{}{\runningpagewiselinenumbers}}
\makeatother

\usepackage{times}
\usepackage{latexsym}

\usepackage[T1]{fontenc}
\usepackage[utf8]{inputenc}

\usepackage{microtype}

\usepackage{inconsolata}

\usepackage{graphicx}

\usepackage{amsmath}
\usepackage{amssymb}
\usepackage{booktabs}
\usepackage{xcolor}

\usepackage{amsthm}
\newtheorem{proposition}{Proposition}
\newtheorem{remark}{Remark}

\usepackage{algorithm}
\usepackage{algpseudocode}

\usepackage{newfloat}
\usepackage{listings}
\DeclareCaptionStyle{ruled}{labelfont=normalfont,labelsep=colon,strut=off}
\floatstyle{ruled}
\newfloat{listing}{tb}{lst}
\floatname{listing}{Listing}

\usepackage{placeins}

\usepackage{array}
\definecolor{deltaGain}{HTML}{009E73}
\definecolor{deltaLoss}{HTML}{D55E00}

\title{SCARL: Sequence Confidence-Aligned Reinforcement Learning}

\author{Zhenghong Ke\textsuperscript{1,2}\thanks{Work done when
          Zhenghong Ke was an intern at Baidu.}, \quad
        Jingbo Zhou\textsuperscript{1}\thanks{Corresponding authors.}, \quad
        Hua Wu\textsuperscript{1}\footnotemark[2] \\
        {\normalsize\mdseries
        \textsuperscript{1}Baidu Frontier Research Department \quad
        \textsuperscript{2}Nanjing University} \\
        {\normalsize\mdseries\texttt{kezhh163@163.com},
        \texttt{\{zhoujingbo, wu\_hua\}@baidu.com}}}

\begin{document}
\maketitle

% scarl_abstract.tex: abstract.

\begin{abstract}
Reinforcement learning with verifiable rewards (RLVR) improves
language-model reasoning by scoring sampled responses with a
rule-based verifier and reinforcing those that pass. Where a
classifier's training couples predicted probability to labels,
nothing ties an RLVR policy's confidence in a response to its
reward: aligned training would drive responses with the same
advantage toward equal confidence, yet the advantage is blind to
confidence.
Under this \emph{sequence-confidence misalignment}, a success the
policy barely found earns the same update as one it produces with
near certainty, and a confident error is corrected no harder than an
exploratory one. We propose SCARL
(\emph{Sequence Confidence-Aligned Reinforcement Learning}), which
reshapes, but never re-signs, each advantage with the alignment gap
between the response's group-relative confidence and the outcome
its advantage certifies, normalized into shares of each class's
advantage budget. Misaligned samples absorb the force, responses with the same
advantage are driven toward equal confidence, and repairing the
misalignment doubles as a remedy for the entropy collapse that
erodes exploration. SCARL provably reduces to its base estimator
when a class's confidences agree, adds no extra passes and one
hyperparameter. On Qwen3-4B-Base it lifts average Pass@32 by
$2.4$ points over DAPO and is best among nine compute-matched
methods on both coverage and per-draw accuracy; applied unchanged
across four estimators and four backbones, it improves $14$ of $16$
settings while preserving policy entropy and widening the
confidence--correctness separation. Our code is available at
\url{https://github.com/kezhh163/scarl}.
\end{abstract}

% scarl_intro.tex: introduction and Figure 1 (fig/teaser.pdf).

\section{Introduction}
\label{sec:intro}

Reinforcement learning with verifiable rewards (RLVR) has become the
standard recipe for turning base language models into reasoners:
sample responses, check them with a rule-based verifier, and
reinforce what passes~\citep{lambert2024tulu, guo2025deepseekr1}.
Group-relative estimators in the GRPO
family~\citep{shao2024deepseekmath} made the recipe practical at
scale by replacing the value model with within-group reward
normalization, and a line of successors refines its clipping,
ratios, sampling, and normalizations~\citep{yu2025dapo,
zheng2025gspo, sapo2025, liu2025drgrpo}.

The recipe also has a well-documented failure mode. Policy entropy
collapses early in training~\citep{yu2025dapo}, exploration narrows,
and the gains concentrate in per-draw accuracy while large-$k$
coverage stalls: RLVR often sharpens the sampling distribution over
problems the model could already solve rather than enlarging that
set~\citep{yue2025rlvr}; entropy falls when log-probability and
advantage covary positively~\citep{cui2025entropy}. Existing
responses widen the upper clipping range~\citep{yu2025dapo},
regulate that covariance~\citep{cui2025entropy, wang2026cwgrpo}, or
refine the update pathway for stability~\citep{sapo2025,
zheng2025gspo}; a growing line conditions the update on the
policy's own confidence or uncertainty, measured per token,
response, or question~\citep{chen2025seedgrpo, cheng2025reasoning,
xie2025ucas, dai2025cde, li2026progrpo, xu2026ace, wang2026capo}.

\begin{figure*}[t]
\centering
\includegraphics[width=0.52\textwidth]{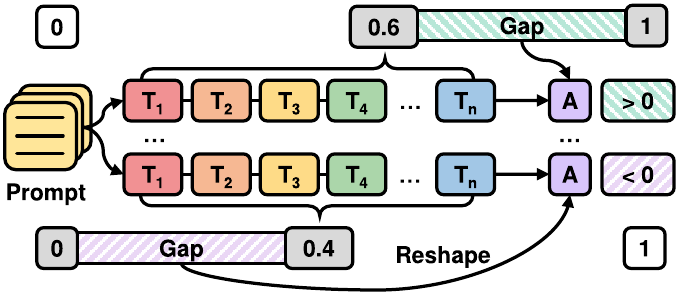}
\caption{\textbf{Sequence-confidence misalignment and its repair.}
Group-relative RLVR gives each response a sequence-level advantage
($A$) blind to how confidently the policy produced it. SCARL turns
token log-probabilities into a group-relative sequence confidence
(gray; here $0.6$ and $0.4$), reads the advantage sign as the
outcome to match ($1$ or $0$), and reshapes $A$ with their distance,
the \emph{alignment gap} (hatched): misaligned responses absorb the
update, aligned ones are left alone.}
\label{fig:teaser}
\end{figure*}

We start from what confidence-aware training should achieve. In a
well-trained classifier, confidence is aligned with labels, the
product of a loss that pushes hardest exactly where belief and label
disagree. The group-relative advantage has no counterpart at the
level where the reward lives: it is a function of the group's
rewards alone, and how confidently the policy produced a response,
its sequence log-likelihood, enters nowhere. A correct response the
policy barely found thus earns the same gradient as one it produces
with near certainty, and a confident error is corrected no harder
than a tentative, exploratory one. We call this
\emph{sequence-confidence misalignment} and state the criterion that
aligned training would satisfy: responses earning the same outcome
should be held with the same confidence. The standard update
violates it in both outcome classes.

We propose SCARL (\emph{Sequence Confidence-Aligned Reinforcement
Learning}), which repairs this misalignment with a single
multiplicative shaping factor on the advantage, its \emph{alignment
gap} (Figure~\ref{fig:teaser}): the distance between each
response's group-relative confidence, computed from the
log-likelihood already materialized for the importance ratios, and
the binarized outcome it earned. Misaligned responses absorb the
force while aligned ones are left alone: the gap damps mastered
successes, consolidates fragile ones, amplifies the correction of
confident errors, and tolerates exploratory ones. Confidences of
responses with the same advantage therefore converge, their
generation probabilities equalize, and the policy keeps producing
diverse solutions, so repairing the misalignment doubles as a remedy
for entropy collapse. Unlike an additive confidence bonus, the
factor is multiplicative and provably positive for any temperature,
so no update is ever reversed; normalized within each outcome
class, it preserves the class's advantage budget and reduces exactly
to the base estimator when a class's confidences agree. SCARL adds
no extra passes, one hyperparameter, and drops into any
group-relative estimator unchanged.

Our contributions are threefold: (i)~we state
\emph{sequence-confidence alignment} as an operational criterion for
group-relative RLVR, same outcome, same confidence, show that the
standard advantage violates it in both outcome classes by
construction, and link the violation to entropy collapse; (ii)~we
introduce SCARL, an alignment-gap reshaping of the advantage that
provably never reverses an update, reduces exactly to the base
estimator when a class's confidences agree, and drops into any
group-relative surrogate with one hyperparameter
(\S\ref{sec:method}); and (iii)~we validate it under a
compute-matched protocol (\S\ref{sec:experiments}): on Qwen3-4B-Base,
SCARL lifts DAPO's average Pass@32 by $2.4$ points ($69.5$ vs.\
$67.1$) and attains the best coverage and per-draw accuracy among
nine methods; applied unchanged across four estimators and four
backbones, it improves $14$ of $16$ pairings; policy entropy stays
above every base estimator's from step $12$ onward and the
confidence separation rises into the large-effect regime; and
ablations show that class-wise normalization is load-bearing
($-9.4$ without it), that outcome-blind shaping forfeits the gain,
and that the whole-response confidence outperforms its per-token
mean.

% scarl_method.tex: method, including the Background subsection
% (labels sec:prelim, eq:grpoadv, eq:ratio, eq:grpo live here).

\section{SCARL: Sequence Confidence-Aligned Reinforcement Learning}
\label{sec:method}

SCARL augments any group-relative policy-gradient algorithm with a
single multiplicative shaping factor on the advantage, built to
repair the sequence-confidence misalignment of \S\ref{sec:intro}.
Uniform advantages violate the alignment criterion on both outcome
classes, and we argue this misallocation is one driver of entropy
collapse in GRPO training~\citep{yu2025dapo, shao2024deepseekmath}:
high-confidence sequences receive disproportionate reinforcement and
the policy sharpens prematurely. The converse is this paper's
constructive claim: when responses with the same advantage are
driven toward equal confidence, their generation probabilities
equalize and the policy retains diverse solutions
(\S\ref{sec:mechanism}).

% ---------------------------------------------------------------------
\subsection{Background: Group-Relative RLVR}
\label{sec:prelim}

An autoregressive policy $\pi_\theta$ maps a prompt
$x \sim \mathcal D$ to a response $y$ with
$\pi_\theta(y \mid x) = \prod_{t}
\pi_\theta(y_t \mid x, y_{<t})$. In
RLVR~\citep{lambert2024tulu, guo2025deepseekr1} the reward is a
rule-based verifier, here $R(x, y) = +1$ if the final answer is
accepted and $-1$ otherwise.
GRPO~\citep{shao2024deepseekmath} dispenses with a value model: for
each prompt it samples a group of $G$ responses
$\{y_i\}_{i=1}^{G} \sim \pi_{\theta_{\mathrm{old}}}(\cdot \mid x)$,
scores them $r_i = R(x, y_i)$, and normalizes rewards within the
group,
\begin{equation}
\hat A_{i,t} \;=\; \hat A_i \;=\;
\frac{r_i - \mathrm{mean}\!\left(\{r_j\}_{j=1}^{G}\right)}
{\mathrm{std}\!\left(\{r_j\}_{j=1}^{G}\right)} ,
\label{eq:grpoadv}
\end{equation}
broadcasting the sequence-level advantage to every token of $y_i$;
Eq.~\eqref{eq:grpoadv} depends on the rewards alone, and the
policy's confidence in $y_i$ enters nowhere. With the token-level
importance ratio
\begin{equation}
\rho_{i,t}(\theta) \;=\;
\frac{\pi_{\theta}\!\left(y_{i,t} \mid x,\, y_{i,<t}\right)}
{\pi_{\theta_{\mathrm{old}}}\!\left(y_{i,t} \mid x,\, y_{i,<t}\right)} ,
\label{eq:ratio}
\end{equation}
GRPO maximizes the clipped surrogate
\begin{equation}
\begin{aligned}
\mathcal{J}_{\textsc{grpo}}(\theta)
={}& \mathbb{E}\Bigg[
\frac{1}{G}\sum_{i=1}^{G}\frac{1}{|y_i|}\sum_{t=1}^{|y_i|}
\min\!\Big(\rho_{i,t}(\theta)\, \hat A_{i,t},\\
&\quad
\mathrm{clip}\!\big(\rho_{i,t}(\theta),\, 1-\varepsilon,\,
1+\varepsilon\big)\, \hat A_{i,t}\Big)\Bigg],
\end{aligned}
\label{eq:grpo}
\end{equation}
over $x$ and $\{y_i\} \sim \pi_{\theta_{\mathrm{old}}}$, optionally
with a KL penalty toward a frozen reference, following each
estimator's published choice. GRPO's descendants (DAPO, GSPO, SAPO)
retain group-relative advantages and vary the surrogate around them;
we write $\mathcal{J}_{\mathrm{base}}\big(\theta; \{\hat
A_{i,t}\}\big)$ for any such objective. SCARL operates exclusively
on $\{\hat A_{i,t}\}$ and is agnostic to these surrogate choices.

\paragraph{The SCARL transformation.}
SCARL leaves rewards, importance ratios, and clipping untouched, and
reshapes the base advantage with a strictly positive sequence-level
shaping factor derived from the policy's own confidence:
\begin{equation}
\hat A^{\textsc{scarl}}_{i,t} \;=\; w_i \,\hat A_{i,t},
\qquad w_i > 0 ,
\label{eq:scarl}
\end{equation}
where $w_i$ is produced in four steps: (i)~read off each sequence's
log-likelihood under the behavior policy (\S\ref{sec:conf}),
(ii)~convert it into a group-relative confidence
(\S\ref{sec:relconf}), (iii)~take the \emph{alignment gap} between
that confidence and the binarized outcome (\S\ref{sec:signaware}),
and (iv)~normalize within each target class (\S\ref{sec:classnorm}).
Every step is a group-level operation on scalars already
materialized during training; the one substantive hyperparameter is
the temperature $\tau$.

% ---------------------------------------------------------------------
\subsection{Sequence Confidence}
\label{sec:conf}

For each response $y_i$ in a group, we define its confidence as the
sequence log-likelihood under the behavior policy,
\begin{equation}
\begin{aligned}
c_i &= \sum_{t=1}^{|y_i|} \log \pi_{\theta_{\mathrm{old}}}
\!\left(y_{i,t} \,\middle|\, x,\, y_{i,<t}\right)\\
&= \log \pi_{\theta_{\mathrm{old}}}(y_i \mid x) \le 0 .
\end{aligned}
\label{eq:conf}
\end{equation}
Two properties motivate this choice. First, $c_i$ reflects the
model's beliefs at sampling time and remains \emph{constant} across
the minibatch updates of an epoch, so $w_i$ is a fixed coefficient
rather than a moving target. Second, the per-token log-probabilities
are already materialized for the importance ratios, so $c_i$ costs a
single summation. Because $c_i$ sums over tokens, longer responses
are, all else equal, judged less confident. We retain the sum
deliberately: it measures the probability the policy committed to
the whole response, and within a group every response answers the
same prompt, so length differences reflect behavioral choices rather
than nuisance variation (\S\ref{sec:ablation} tests the per-token
alternative).

% ---------------------------------------------------------------------
\subsection{Group-Relative Confidence}
\label{sec:relconf}

Raw log-likelihoods are not comparable across prompts, and their
dispersion varies by orders of magnitude across models and training
stages, so we standardize within the group before relativizing. Let
$\sigma_{\mathcal G} = \mathrm{std}\!\left(\{c_j\}_{j=1}^{G}\right)$,
$\tau > 0$ a temperature, and $\delta > 0$ a small stabilizer; then
\begin{equation}
\tilde c_i \;=\; \frac{c_i}{\tau \left(\sigma_{\mathcal G} +
\delta\right)},
\qquad
p_i \;=\; \frac{\exp(\tilde c_i)}{\sum_{j=1}^{G} \exp(\tilde c_j)} .
\label{eq:softmax}
\end{equation}
Three properties matter. \emph{Shift invariance:} any offset common
to the group (intrinsic prompt difficulty) cancels in the softmax,
so $p_i$ depends only on how confident the policy was in $y_i$
relative to its siblings, mirroring the base estimator's reward
normalization. \emph{Calibrated sharpness:} the scaled logits have
standard deviation $1/\tau$ (up to $\delta$), decoupling softmax
concentration from group dispersion, so a single $\tau$ transfers
across prompts, models, and training stages (\S\ref{sec:ablation}).
\emph{Positivity:} $p_i \in (0,1)$ with $\sum_i p_i = 1$, which is
what guarantees that shaping can rescale an advantage but never flip
its sign.

% ---------------------------------------------------------------------
\subsection{Alignment-Gap Shaping}
\label{sec:signaware}
% Label kept for cross-file stability.

We call confidence \emph{aligned} on a group when responses with the
same outcome are held with the same confidence; the alignment
criterion asks how far each sequence stands from that state. Write
$t_i = \mathbf{1}\{\hat A_i \ge 0\}$ for the binarized outcome that
sample $i$'s advantage certifies; the classes follow the advantage
sign, never the raw reward (under our binary reward the two
coincide; groups with identical rewards have $\hat A_i = 0$ and are
inert). We scale each sample's update by its \emph{alignment gap}
(Figure~\ref{fig:teaser}),
\begin{equation}
\tilde w_i \;=\; \bigl|\, t_i - p_i \,\bigr| \;=\;
\begin{cases}
1 - p_i & \text{if } t_i = 1 ,\\[2pt]
\;p_i & \text{if } t_i = 0 .
\end{cases}
\label{eq:signaware}
\end{equation}
The outcome decides only the \emph{direction} of an update;
the gap decides \emph{how much} of it survives, and it orients the
update within each class. Aligned samples are left nearly untouched:
the mastered success and the tentative, exploratory error both carry
a small gap. Misaligned samples absorb the force: a fragile success
the policy barely found is consolidated, and an error committed with
conviction is corrected hardest. In form, the gap is focal loss's
modulating factor~\citep{lin2017focal} with the group share as
prediction and the certified outcome as label;
\S\ref{sec:classnorm} normalizes it within each class.

% ---------------------------------------------------------------------
\subsection{Class-Wise Normalization}
\label{sec:classnorm}

The two branches of Eq.~\eqref{eq:signaware} cannot be used as-is:
since $\sum_i p_i = 1$, near-uniform confidence gives samples with
$t_i = 1$ gaps near $(G-1)/G$ but samples with $t_i = 0$ gaps near
$1/G$, an imbalance that all but erases penalties. We therefore
normalize within target classes
$\mathcal S_{+} = \{ i : t_i = 1 \}$ and
$\mathcal S_{-} = \{ i : t_i = 0 \}$, with $\mathcal S(i)$ the class
containing $i$:
\begin{equation}
w_i \;=\;
\frac{\tilde w_i}{\dfrac{1}{|\mathcal S(i)|}
\sum_{j \in \mathcal S(i)} \tilde w_j } ,
\label{eq:classnorm}
\end{equation}
with $w_i = 1$ whenever $|\mathcal S(i)| \le 1$. The division is a
conservation law: Eq.~\eqref{eq:classnorm} converts the raw gaps
into \emph{shares} of each class's fixed advantage budget
($\mathbb{E}[\, w_i \mid \mathcal S(i)\,] = 1$), so confidence
decides how the budget is spent, never how large it is. It also
measures each response against its own class: within each class,
Eq.~\eqref{eq:classnorm} makes a response's update proportional to
how far its confidence lies from the class mean: a success held
with less confidence than its correct siblings is consolidated, and
an error held with more confidence than its incorrect siblings is
corrected hardest (the appendix gives the algebra). Two guarantees
follow (proofs in the appendix). \emph{Property 1 (sign preservation):} every $w_i$ is
strictly positive and bounded by $|\mathcal S(i)|$, so
Eq.~\eqref{eq:scarl} never reverses the direction of improvement.
\emph{Property 2 (class-mass preservation and recovery):} when all
responses in a class share the same group-relative confidence,
$w_i = 1$ exactly and SCARL coincides with the base estimator: it is
a strict generalization of GRPO that activates only where confidence
varies within an outcome class.

% ---------------------------------------------------------------------
\subsection{Training Objective}
\label{sec:objective}

SCARL specifies only the advantage transformation; the surrogate is
inherited unchanged. Given any group-relative objective
$\mathcal{J}_{\mathrm{base}}\big(\theta; \{\hat A_{i,t}\}\big)$, the
SCARL objective is
$\mathcal{J}_{\mathrm{base}}\big(\theta; \{w_i \hat A_{i,t}\}\big)$,
instantiated throughout on GRPO's clipped surrogate
(Eq.~\eqref{eq:grpo}). Since $w_i$ is a function of
$\pi_{\theta_{\mathrm{old}}}$ only, no gradient flows through it.
Intervening purely in advantage space, SCARL composes with ratio-space
interventions~\citep{yu2025dapo, sapo2025, zheng2025gspo}, dynamic
sampling, and alternative base
estimators~\citep{ahmadian2024rloo, liu2025drgrpo} rather than
competing with them; an algorithm and a reference implementation
appear in the appendix.

\paragraph{Why it helps (informal).}
\emph{Property 3 (entropy-aware shaping):} among positive-advantage
samples $w_i$ is anti-monotone in confidence, damping precisely the
high-$\log\pi$, high-$A$ terms that dominate the entropy-decay
covariance~\citep{cui2025entropy}; among negative-advantage samples
it concentrates correction on over-confident modes and spares rare
exploratory errors. Both effects slow entropy decay at its source
(\S\ref{sec:mechanism}; appendix). \emph{Property 4 (cost):}
$O(G)$ scalar operations per group on quantities already computed
for the surrogate; wall-clock overhead is negligible.

% scarl_experiments_setup.tex: experiments overview and setup. The
% full configuration lives in the supplement (tab:hparams); the
% values quoted here are the narrative subset.

\section{Experiments}
\label{sec:experiments}

Our experiments ask whether SCARL outperforms strong RLVR baselines
under a compute-matched recipe (\S\ref{sec:main}), whether the
improvement transfers across estimators and model families without
retuning (\S\ref{sec:generality}), whether the gain comes from
alignment itself (\S\ref{sec:ablation}), and whether SCARL acts
through its intended mechanism (\S\ref{sec:mechanism}).

\subsection{Setup}
\label{sec:setup}

\paragraph{Models, estimators, and baselines.}
The primary comparison uses \textbf{Qwen3-4B-Base}; the generality
study adds Qwen3-1.7B-Base, DeepSeek-R1-Distill-Qwen-1.5B, and
Hunyuan-1.8B-Pretrain. We evaluate SCARL as a drop-in modifier of
four base estimators: GRPO~\citep{shao2024deepseekmath},
DAPO~\citep{yu2025dapo}, GSPO~\citep{zheng2025gspo}, and
SAPO~\citep{sapo2025}. The main comparison adds
Dr.~GRPO~\citep{liu2025drgrpo} and three confidence-aware methods
run in their native configurations:
Entropy Advantage Shaping (EAS)~\citep{cheng2025reasoning},
SEED-GRPO~\citep{chen2025seedgrpo}, and UCAS~\citep{xie2025ucas}.

\paragraph{Benchmarks and metrics.}
We evaluate on five competition-mathematics benchmarks: AMC23,
AIME24, AIME25, Minerva Math~\citep{lewkowycz2022minerva}, and
MATH-500~\citep{hendrycks2021math, lightman2024letsverify}, sampling
$k = 32$ responses per problem. \emph{Average accuracy}
($\mathrm{Avg}@k = c/k$ for $c$ correct responses) measures sampling
efficiency; \emph{coverage} ($\mathrm{Pass}@k$) measures whether at
least one response is correct. A method can raise the former while
leaving the latter unchanged, sharpening the distribution over an
unchanged solvable set~\citep{yue2025rlvr}, so we track both:
Pass@32 carries every table and Avg@32 the training curves of the
main comparison; Pass@32 is estimated by bootstrap resampling of
each problem's $32$ responses (Appendix~\ref{app:hparams}).

\paragraph{Confidence separation (Cohen's $d$).}
We measure alignment with the confidence signal $c_i$ of
Eq.~\eqref{eq:conf} itself. Within each group $g$ we split rollouts
into correct ($\hat A_i \ge 0$) and incorrect ($\hat A_i < 0$) sets
with confidence means $\mu^{\pm}_g$ and variances $s^{2,\pm}_g$;
pooling across a step's groups,
\begin{equation}
d \;=\; \frac{\mathrm{mean}_g\!\left(\mu^{+}_g - \mu^{-}_g\right)}
{\sqrt{\big(V^{+} + V^{-}\big)/2}},
\quad
V^{\pm} = \mathrm{mean}_g\, s^{2,\pm}_g .
\label{eq:cohend}
\end{equation}
A larger $d$ means the policy's confidence better distinguishes its
correct from its incorrect responses; $d$ rises both when the
classes separate and when same-class confidences converge.

\paragraph{Training and evaluation protocol.}
All methods share one recipe and differ only in the advantage
transformation (SCARL) or the surrogate (base estimators), so
comparisons are compute-matched: $150$ steps, group size $G = 8$,
batch size $512$, learning rate $1\times10^{-6}$, trained on
DAPO-Math-17K~\citep{yu2025dapo} with
verl~\citep{sheng2024hybridflow}; SCARL uses $\tau = 5$ throughout,
selected once by the sweep in \S\ref{sec:ablation}. Evaluation draws
$32$ responses per problem (temperature $0.6$, top-$p$ $0.95$,
top-$k$ $20$) every $5$ steps; tables report each run at its best
checkpoint by five-benchmark average, and correctness is checked by
rule-based verifiers matched to each data source. Full
hyperparameters, per-estimator settings, verifier routing, and
prompts appear in the appendix.

% scarl_experiments_main.tex: main results. Figure regenerated by
% plot_main_curves_compact.py.

\subsection{Main Results: SCARL versus Existing Interventions}
\label{sec:main}

We apply SCARL to DAPO, denoted \emph{DAPO\,+\,SCARL}. On
Qwen3-4B-Base it improves average Pass@32 by $2.4$ points over
vanilla DAPO ($69.5$ vs.\ $67.1$), the controlled same-base measure
of SCARL's effect, and attains the best average Pass@32 and Avg@32
among all nine methods: improving both means the policy solves more
problems within the $32$-sample budget rather than sharpening its
distribution over an unchanged solvable set. Every method improves
on the untrained base model's $54.5$ average Pass@32; SCARL improves
it the most ($+15.0$). The advantage holds on Qwen3-1.7B-Base, where
DAPO\,+\,SCARL again takes the highest average ($58.0$, $+2.2$ over
DAPO; table in the appendix). Figure~\ref{fig:main_curves} shows the
training curves and Table~\ref{tab:main_pass32} the per-benchmark
breakdown.

\begin{figure*}[t]
\centering
\includegraphics[width=\linewidth]{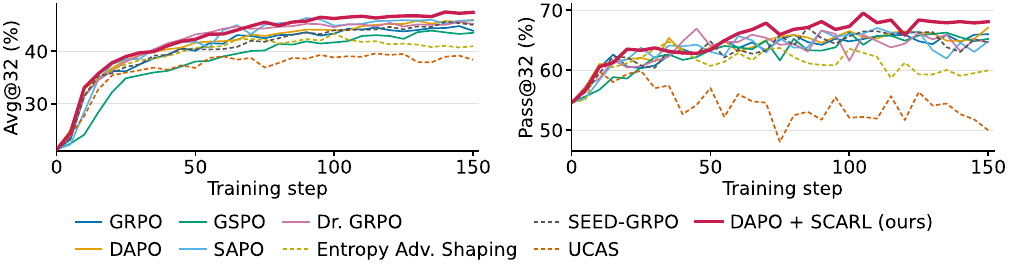}
\caption{Avg@32 (left) and Pass@32 (right) on Qwen3-4B-Base,
averaged over the five benchmarks, across $150$ training steps; step
$0$ is the untrained base model. SCARL (bold) leads both metrics
through the second half of training, with a widening margin.}
\label{fig:main_curves}
\end{figure*}

\paragraph{Training dynamics.}
SCARL is the top method at $20$ of the $30$ logged steps on both
metrics, and its lead over the next-best method grows over the
second half of training ($1.1$ points on Avg@32 and $1.6$ on Pass@32
over the final ten checkpoints).

\paragraph{Per-benchmark results.}
SCARL is best on four of the five benchmarks and trails only on
Minerva, where Dr.~GRPO leads ($59.6$ vs.\ $57.4$); the largest
margins fall on the AIME sets.

\paragraph{Confidence-awareness alone is not enough.}
The confidence-aware baselines behave very differently from one
another: SEED-GRPO is competitive ($67.0$),
Entropy Advantage Shaping is no better than vanilla GRPO ($64.7$
vs.\ $66.0$), and UCAS falls well below it ($59.9$). Coupling the
update to the policy's (un)certainty is not automatically
beneficial; the gain lies in the alignment-gap construction.

% tab_main_pass32.tex: main comparison table. Source:
% main_result_pass@32_Qwen3-4B.csv (note: the CSV lists AIME25
% before AIME24; this table uses the conventional order).
\begin{table*}[t]
\centering
\caption{\textbf{Main comparison on Qwen3-4B-Base.} Pass@32 on five
competition-mathematics benchmarks (\%), at each method's best
checkpoint by average Pass@32. \textbf{Bold} marks the best value
per column; the untrained base model is a reference, excluded from
bolding.}
\label{tab:main_pass32}
\footnotesize
\setlength{\tabcolsep}{6pt}
\begin{tabular}{@{}lcccccc@{}}
\toprule
Method & AMC23 & AIME24 & AIME25 & Minerva & MATH-500 & Avg. \\
\midrule
Base model (no RL) & 75.1 & 25.9 & 23.3 & 56.3 & 92.0 & 54.5 \\
\addlinespace[2pt]
\multicolumn{7}{@{}l}{\itshape Base estimators}\\
GRPO & 88.9 & 44.6 & 44.7 & 56.4 & 95.3 & 66.0 \\
DAPO & 90.4 & 46.1 & 45.0 & 58.4 & 95.5 & 67.1 \\
GSPO & 91.8 & 45.3 & 42.9 & 58.3 & 92.8 & 66.2 \\
SAPO & 92.2 & 45.7 & 43.1 & 59.3 & 94.8 & 67.0 \\
Dr. GRPO & 88.9 & 47.9 & 43.3 & \textbf{59.6} & 94.9 & 66.9 \\
\addlinespace[2pt]
\multicolumn{7}{@{}l}{\itshape Confidence-aware methods}\\
Entropy Advantage Shaping & 89.2 & 42.0 & 43.9 & 55.8 & 92.7 & 64.7 \\
SEED-GRPO & 89.7 & 45.5 & 48.0 & 58.0 & 94.0 & 67.0 \\
UCAS & 85.8 & 33.5 & 30.3 & 58.0 & 91.6 & 59.9 \\
\addlinespace[2pt]
DAPO\,+\,SCARL (ours) & \textbf{95.3} & \textbf{49.0} & \textbf{50.1} & 57.4 & \textbf{95.6} & \textbf{69.5} \\
\bottomrule
\end{tabular}
\end{table*}

% scarl_experiments_generality.tex: generality study. Full
% per-benchmark table in the supplement.

\subsection{Generality across Base Estimators and Model Families}
\label{sec:generality}

SCARL shapes only the advantage (\S\ref{sec:objective}), so it
should compose with any group-relative estimator; we apply it
unchanged, with $\tau = 5$, to four estimators on four backbones
from three model families.
Table~\ref{tab:generality} reports all sixteen pairings: SCARL
improves average Pass@32 in $14$ of the $16$ (mean $\Delta$ of
$+1.0$), and on every benchmark it improves a majority of the
pairings (per-benchmark tables in the appendix). A single
confidence-aware factor thus transfers across estimators and model
families without per-setting retuning, and every one of the $32$
trained runs raises coverage over its untrained backbone.

% tab_generality.tex: generality summary. Source:
% Pass@32_multimodel_multialgos.csv; deltas computed from unrounded
% averages, then rounded. Full paired base/+SCARL per-benchmark
% table: supplement.
\definecolor{deltaGain}{HTML}{009E73}
\definecolor{deltaLoss}{HTML}{D55E00}
\begin{table}[t]
\centering
\caption{\textbf{SCARL as a drop-in modifier.} Average Pass@32 (\%)
of each estimator with SCARL added, on four backbones (in
parentheses: change vs.\ the plain estimator; per-benchmark values in
the appendix). DS-R1-1.5B: DeepSeek-R1-Distill-Qwen-1.5B.}
\label{tab:generality}
\small
\setlength{\tabcolsep}{2pt}
\resizebox{\columnwidth}{!}{%
\begin{tabular}{@{}lcccc@{}}
\toprule
Backbone & +GRPO & +DAPO & +GSPO & +SAPO \\
\midrule
Qwen3-4B-Base &
67.3(\textcolor{deltaGain}{+1.4}) &
69.5(\textcolor{deltaGain}{+2.4}) &
67.7(\textcolor{deltaGain}{+1.5}) &
68.1(\textcolor{deltaGain}{+1.1}) \\
Qwen3-1.7B-Base &
56.3(\textcolor{deltaGain}{+1.8}) &
58.0(\textcolor{deltaGain}{+2.2}) &
57.1(\textcolor{deltaGain}{+0.7}) &
55.6(\textcolor{deltaLoss}{$-$0.2}) \\
DS-R1-1.5B &
70.5(\textcolor{deltaGain}{+1.2}) &
72.0(\textcolor{deltaGain}{+0.5}) &
71.4(\textcolor{deltaGain}{+0.5}) &
71.6(\textcolor{deltaGain}{+0.1}) \\
Hunyuan-1.8B &
62.0(\textcolor{deltaGain}{+0.8}) &
64.4(\textcolor{deltaGain}{+1.5}) &
62.1(\textcolor{deltaGain}{+0.1}) &
62.5(\textcolor{deltaLoss}{$-$0.2}) \\
\bottomrule
\end{tabular}}
\end{table}

\paragraph{The two exceptions.}
Both pairings in which \mbox{+\,SCARL} does not improve are under
SAPO, and both are small ($-0.2$ each). SAPO is the one base
estimator that already rescales every update continuously, through
its temperature-controlled gate in ratio space, so stacking a second
continuous modulation yields the smallest marginal benefit; the
other three estimators gate updates only through hard clipping and
improve with SCARL in all twelve settings.

% scarl_experiments_ablation.tex: ablations and temperature sweep.

\subsection{Ablations: Alignment Is the Active Ingredient}
\label{sec:ablation}

SCARL has two structural components, one confidence signal, and one
substantive hyperparameter: alignment-gap shaping
(\S\ref{sec:signaware}), class-wise normalization
(\S\ref{sec:classnorm}), the summed log-likelihood of
Eq.~\eqref{eq:conf}, and the temperature $\tau$. We ablate each on
Qwen3-1.7B-Base with GSPO as the base estimator, a second backbone
and estimator relative to the main comparison; each variant changes
exactly one element (Table~\ref{tab:ablation}).

% tab_ablation.tex: ablation and temperature summary. Sources:
% ablation_Pass@32_Qwen3-1.7B_GSPO-SCARL.csv,
% tau_with_Qwen3-1.7B_GSPO-SCARL.csv; deltas from unrounded averages.
\definecolor{deltaGain}{HTML}{009E73}
\definecolor{deltaLoss}{HTML}{D55E00}
\begin{table}[t]
\centering
\caption{\textbf{Component ablations and temperature sensitivity}
(Qwen3-1.7B-Base, GSPO base; average Pass@32 over the five
benchmarks). $\Delta$ is relative to plain GSPO; per-benchmark
values in the appendix.}
\label{tab:ablation}
\small
\setlength{\tabcolsep}{5pt}
\begin{tabular}{@{}lcc@{}}
\toprule
Variant & Avg. & $\Delta$ \\
\midrule
GSPO (base estimator) & 56.4 & --- \\
\quad + SCARL w/o class-wise norm. & 47.0 & \textcolor{deltaLoss}{$\mathbf{-9.4}$} \\
\quad + SCARL w/o alignment target & 56.6 & \textcolor{deltaGain}{$+0.2$} \\
\quad + SCARL w/ per-token mean confidence & 55.3 & \textcolor{deltaLoss}{$-1.1$} \\
\quad + SCARL (full; $\tau = 5$) & 57.1 & \textcolor{deltaGain}{$\mathbf{+0.7}$} \\
\addlinespace[2pt]
\multicolumn{3}{@{}l}{\itshape Temperature sweep (full SCARL)}\\
\quad $\tau = 0.2$ & 55.1 & \textcolor{deltaLoss}{$-1.3$} \\
\quad $\tau = 0.5$ & 55.6 & \textcolor{deltaLoss}{$-0.8$} \\
\quad $\tau = 2$ & 56.2 & \textcolor{deltaLoss}{$-0.2$} \\
\quad $\tau = 10$ & 54.7 & \textcolor{deltaLoss}{$-1.7$} \\
\bottomrule
\end{tabular}
\end{table}

\paragraph{Class-wise normalization is load-bearing.}
Training on the raw gaps of Eq.~\eqref{eq:signaware} without
Eq.~\eqref{eq:classnorm} does not merely lose SCARL's gain; it
forfeits RL's. Average Pass@32 falls to $47.0$: $9.4$ points below
the plain base estimator and within $0.4$ of the \emph{untrained}
backbone, the failure mode \S\ref{sec:classnorm} predicts when the
penalty class's advantage budget is all but erased.

\paragraph{The gain lives in alignment.}
The second variant keeps every ingredient except the alignment
target: it computes the gap to a \emph{constant}
($\tilde w_i = 1 - p_i$ for every sample regardless of outcome,
class normalization kept), so the factor still responds to
confidence but carries no information about where confidence
\emph{should} be. Performance returns to the baseline ($56.6$ vs.\
$56.4$; the full construction reaches $57.1$). The gain therefore
comes from measuring confidence against the verified outcome, not
from confidence shaping per se, mirroring \S\ref{sec:main}.

\paragraph{Whole-response confidence.}
Replacing the summed log-likelihood of Eq.~\eqref{eq:conf} with its
per-token mean, a length-normalized confidence, drops average
Pass@32 to $55.3$, $1.1$ points below the plain base estimator. The
mean rates a long response that wanders to its answer
as confidently as a short one that commits to it, so the gap stops
separating fragile successes from mastered ones.

\paragraph{Temperature.}
Sweeping $\tau \in \{0.2, 0.5, 2, 5, 10\}$ yields average Pass@32 of
$55.1$, $55.6$, $56.2$, $57.1$, and $54.7$: rising monotonically to
the default $\tau = 5$ and dropping beyond it, the two failure
directions the calibrated-sharpness design of \S\ref{sec:relconf}
anticipates. $\tau = 5$ was selected once and reused, without
retuning, in all sixteen pairings of \S\ref{sec:generality}.

% scarl_experiments_mechanism.tex: mechanism study. Figure
% regenerated by plot_mechanism_compact.py; full grid in the
% supplement.

\subsection{Mechanism: Preserved Exploration, Repaired Confidence
Alignment}
\label{sec:mechanism}

\begin{figure*}[t]
\centering
\includegraphics[width=\linewidth]{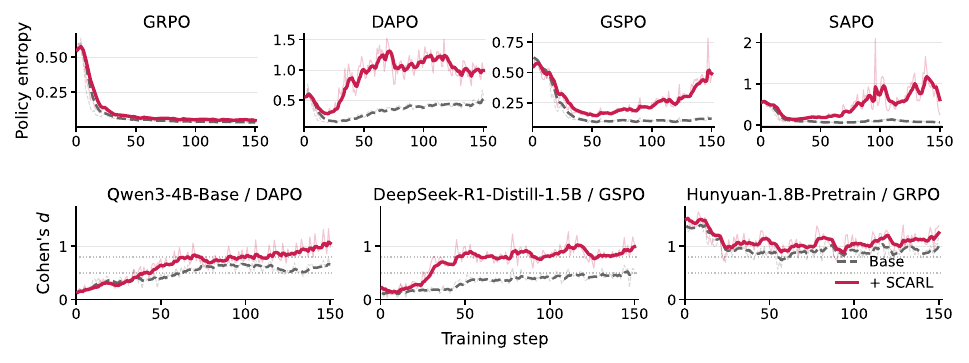}
\caption{\textbf{Mechanism.} \emph{Top:} policy entropy on
Qwen3-4B-Base per base estimator, without (grey dashed) and with
(\textcolor[HTML]{C81D4E}{crimson}) SCARL. \emph{Bottom:} confidence
separation (Cohen's $d$, Eq.~\eqref{eq:cohend}) in three settings;
dotted lines at $d = 0.5$ and $0.8$. EMA-smoothed (span~$8$) over
faint raw values.}
\label{fig:mechanism}
\end{figure*}

If SCARL improves coverage for the reason it was designed, two
signatures should be visible \emph{during} training: entropy should
stay higher than under the base estimator (Property~3), and the
confidence separation of Eq.~\eqref{eq:cohend} should widen.
Figure~\ref{fig:mechanism} shows both, for every base estimator on
Qwen3-4B-Base and for three settings spanning three model families.

\paragraph{Preserved exploration.}
Adding SCARL keeps policy entropy above the base estimator's at
every logged step from step $12$ onward, for all four estimators;
over the final ten steps, mean entropy with SCARL is $2.0\times$ the
base level on DAPO ($1.02$ vs.\ $0.50$), $4.1\times$ on GSPO, and
$11\times$ on SAPO. GRPO is the instructive exception: SCARL slows
the decay at every step but, acting in advantage space alone, cannot
avert collapse under GRPO's symmetric hard clipping, which most
tightly restricts the up-weighting of low-probability
tokens~\citep{yu2025dapo}. Because exploration collapse is how RLVR
narrows the Pass@$k$ boundary~\citep{yue2025rlvr}, preserved entropy
is the proximate explanation for the coverage gains of
\S\ref{sec:main}.

\paragraph{Repaired confidence alignment.}
The policy's own confidence becomes a sharper indicator of
correctness in all three settings. Over the final ten steps, SCARL
lifts Cohen's $d$ from $0.65$ to $1.06$ on Qwen3-4B-Base\,/\,DAPO
and from $0.50$ to $0.97$ on DeepSeek-R1-Distill-Qwen-1.5B\,/\,GSPO,
in both cases alone crossing from the medium into the large-effect
regime ($d > 0.8$); on Hunyuan-1.8B-Pretrain\,/\,GRPO it raises an
already-large $0.95$ to $1.20$. The base run's transiently higher
$d$ over the DAPO setting's first $35$ steps reflects the
confident-success reinforcement that SCARL damps by design
(appendix). Moving both signatures at once is what the construction
intends, and it is why SCARL's gains appear in coverage, not only
per-draw accuracy.

% scarl_related_work.tex: related work.

\section{Related Work}
\label{sec:related}

\paragraph{RL with verifiable rewards.}
RLVR replaces learned reward models~\citep{ouyang2022training} with
programmatic verification~\citep{lambert2024tulu} and, scaled up,
elicits long chain-of-thought reasoning~\citep{guo2025deepseekr1};
it rests on GRPO and its descendants~\citep{shao2024deepseekmath,
yu2025dapo, zheng2025gspo, sapo2025, liu2025drgrpo};
AAPO~\citep{xiong2026aapo} augments the group-relative advantage with
a reward margin over the reference model. Our coverage-centred
protocol follows \citet{yue2025rlvr}.

\paragraph{Entropy collapse and the update pathway.}
Entropy collapse in RLVR is well documented~\citep{yu2025dapo}; for
softmax policies, entropy falls with the covariance between
log-probability and advantage~\citep{cui2025entropy}. Interventions
act at different points of the update: DAPO widens the upper clipping range,
\citet{cui2025entropy} regulate the covariance through bonuses or
detached tokens, CW-GRPO~\citep{wang2026cwgrpo} down-weights tokens
with extreme covariance, and GSPO's sequence-level ratios and SAPO's
smooth gate refine the ratio pathway. All regulate how much of an
update is transmitted; SCARL changes how reinforcement is allocated
across a group's sequences and composes with them
(\S\ref{sec:generality}).

\paragraph{Confidence-aware policy optimization.}
Several methods condition the group-relative update on the policy's
own confidence or uncertainty, measured at the token, response, or
question level. CDE~\citep{dai2025cde} adds a clipped,
annealed perplexity bonus to the reward; EAS~\citep{cheng2025reasoning}
adds a clipped, detached token-entropy bonus to the advantage, and
ProGRPO's ARM~\citep{li2026progrpo} a scaled deviation of a
response's confidence from the group mean, equalizing confidence
across correct responses; SEED-GRPO~\citep{chen2025seedgrpo} scales
each question's advantages by the semantic entropy of its answers,
and UCAS~\citep{xie2025ucas} modulates them with a logit-space
self-certainty and a token-level certainty penalty;
CAPO~\citep{wang2026capo} replaces the advantage with a pairwise
logistic-AUC surrogate over mean token log-probability, targeting
calibration; ACE~\citep{xu2026ace} amplifies the penalty on errors
whose likelihood has risen relative to a reference model. SCARL
shares the premise and differs
in three places (Table~\ref{tab:update_rules}): its criterion covers
both outcome classes and measures each response against its own
class's mean confidence; its factor is multiplicative and positive
for any temperature, so no update is reversed and no bound or clip
is needed; and its class-wise mean-one normalization recovers the
base estimator exactly when a class's confidences agree.
Confidence-based reweighting has precedent in focal
loss~\citep{lin2017focal}, GHM~\citep{li2019ghm}, and prioritized
replay~\citep{schaul2016per}.

% scarl_conclusion.tex: conclusion.

\section{Conclusion}
\label{sec:conclusion}

We identified \emph{sequence-confidence misalignment} as a blind
spot of group-relative RLVR: where a classifier's training ties
confidence to labels, nothing ties an RLVR policy's confidence to
the verified outcome. SCARL restores the missing tie by reshaping
each advantage with its alignment gap, computed from quantities
training already materializes. It delivered the best coverage and
per-draw accuracy among nine compute-matched methods, transferred
across four estimators and four backbones without retuning,
preserved exploration, and widened the confidence--correctness
separation.

% scarl_limitations.tex: Limitations section. Required by ACL Rolling
% Review: a section titled exactly "Limitations", placed after the
% conclusion and before the references without a page break. It does
% not count toward the 8-page limit and may not introduce new
% experiments, figures, or analysis
% (https://aclrollingreview.org/cfp, "Limitations (required section)").

\section*{Limitations}

Our experiments cover competition mathematics with binary,
rule-based verifier rewards; other domains and shaped or continuous
rewards are not studied. The backbones span three model families
but only scales up to 4B parameters. The inference-time use of the
aligned confidence that training produces, for abstention or
selection~\citep{tomani2024abstention}, is a natural extension that
we have not explored.
  % required by ARR: after the conclusion, before the references

\bibliography{scarl}

% Appendix (formerly paper_aaai/supplement.tex), after the references per ARR.
\appendix
% scarl_supplement.tex: extended discussion and additional results
% (input by supplement.tex; cross-references into the main paper
% resolve via xr, so compile main.tex first).

% ---------------------------------------------------------------------
\section{Extended Method Discussion}
\label{supp:discussion}

This section preserves the method discussion compressed out of
\S\ref{sec:method} in the main paper.

\paragraph{The covariance fingerprint, in full.}
For softmax policies, the change in policy entropy under a
policy-gradient step is, to first order, the \emph{negative}
covariance between token log-probability and
advantage~\citep{cui2025entropy}; a group-relative advantage is
broadcast across a response's tokens, so the fingerprint aggregates
to the sequence level, and an estimator that weights sequences by
advantage alone lets confident successes (high $\log \pi$, positive
$\hat A$) dominate that covariance, pushing entropy steadily
downward. The misalignment misallocates learning
even before entropy becomes a concern: updates keep reinforcing
behaviors the model has already internalized while
under-consolidating fragile, newly acquired ones, and they punish
exploratory mistakes as severely as confident errors.

\paragraph{Property 3 (entropy-aware shaping, informal), in full.}
SCARL attacks the covariance directly. Among positive-advantage
samples, $w_i$ is anti-monotone in confidence, damping precisely the
high-$\log\pi$, high-$A$ terms that dominate the covariance; among
negative-advantage samples, $w_i$ is monotone in confidence,
strengthening the updates that move probability mass away from
over-concentrated modes while sparing the rare, exploratory errors
whose suppression would only sharpen the distribution further. Both
effects shrink
$\mathrm{Cov}\!\big(\log \pi,\, w \cdot A\big)$ relative to
$\mathrm{Cov}\!\big(\log \pi,\, A\big)$, slowing entropy decay at
its source rather than through post-hoc clipping or entropy bonuses.

\paragraph{The within-class view.}
Class normalization measures the raw gap of Eq.~\eqref{eq:signaware}
against the class mean. With $\bar p_{\mathcal S}$ the mean
confidence of a class, Eq.~\eqref{eq:classnorm} gives
$w_i = 1 + (\bar p_{\mathcal S} - p_i)/(1 - \bar p_{\mathcal S})$ on
$\mathcal S_{+}$ and
$w_i = 1 + (p_i - \bar p_{\mathcal S})/\bar p_{\mathcal S}
= |\mathcal S_{-}|\, q_i$ on $\mathcal S_{-}$, where
$q_i = p_i / \sum_{j \in \mathcal S(i)} p_j$ is the response's share
of its class's confidence: the factor measures each response against
the uniform within-class share and equals one exactly when the class
is uniform. Two consequences follow. First, within each target class
the optimization speed decreases with confidence and increases with
novelty, in the surprisal sense of \S\ref{sec:conf}, so responses in
the same target class are driven toward a shared confidence and a
shared speed, and no single high-confidence sample can dominate its
class's update. Second, standard training sharpens the
sequence-confidence distribution through two channels at once: it
grows the peaks by reinforcing confident successes and erases the
tails by punishing rare exploratory errors. The factor throttles
both channels, keeping the distribution smoother and entropy higher.
This is the distributional restatement of the covariance argument
above, and it is the effect \S\ref{sec:mechanism} measures.

\paragraph{A per-sequence learning-rate reading.}
At the start of each inner epoch ($\theta = \theta_{\mathrm{old}}$,
hence $\rho_{i,t} = 1$), the unclipped update direction is
$\mathbb{E}\big[\, w_i\, \hat A_i\, \nabla_\theta \log
\pi_\theta(y_i \mid x) \big]$: SCARL acts as a per-sequence,
confidence-dependent learning rate on the vanilla policy gradient.

\paragraph{Adaptive-temperature details.}
Without the group standardization of Eq.~\eqref{eq:softmax},
near-degenerate groups ($\sigma_{\mathcal G}\!\to\!0$) yield
uninformative uniform factors while heterogeneous groups collapse
onto a near one-hot vector; with it, a single $\tau$ transfers
across prompts, model scales, and training stages. A
fixed-temperature variant, $\tilde c_i = c_i/\tau$, is possible but
requires retuning $\tau$ per model and training stage.

\paragraph{Reweighting lineage.}
Equation~\eqref{eq:signaware} can be read as a sequence-level
analogue of focal modulation~\citep{lin2017focal}: focal loss
down-weights well-classified examples through a factor that vanishes
as the prediction approaches the label, and
Eq.~\eqref{eq:signaware} is that factor in form, with the certified
outcome as the label and the group share as the prediction,
transported into advantage space; Eq.~\eqref{eq:classnorm} then
measures it against the class mean rather than the label. It
likewise echoes prioritized experience
replay~\citep{schaul2016per}, which allocates more learning to
surprising transitions.

\paragraph{Relation to other interventions.}
Existing interventions on the group-relative update act in different
spaces and with different aims
(Table~\ref{tab:interventions}). Two target entropy collapse
explicitly: DAPO's clip-higher widens the trust region
asymmetrically in \emph{ratio} space, and entropy bonuses or
covariance-based
corrections act on the \emph{regularizer or gradient}. Two are
motivated by training stability rather than entropy: GSPO changes
the \emph{granularity} at which ratios are computed, and SAPO
replaces hard clipping with a smooth sigmoid gate. All four regulate
the \emph{transmission} of an update, that is, how much of it
survives ratios, clipping, or bonuses; the \emph{allocation} of
reinforcement across sequences is inherited unchanged from the
advantage. SCARL is complementary: it embeds the policy's confidence
into the \emph{advantage} itself, so high-confidence correct actions
receive less reinforcement by construction, and the policy sharpens
more slowly at the source, before any clipping or gating is applied.

\begin{table}[t]
\centering
\caption{Where different interventions on the group-relative update
act. SCARL targets sequence-confidence misalignment in advantage
space and composes with all of them.}
\label{tab:interventions}
\footnotesize
\setlength{\tabcolsep}{4pt}
\begin{tabular}{@{}p{0.27\columnwidth}p{0.19\columnwidth}p{0.36\columnwidth}@{}}
\toprule
Method & Space & Mechanism \\
\midrule
Clip-Higher~\citep{yu2025dapo} & importance ratio &
asymmetric clipping range \\
SAPO~\citep{sapo2025} & importance ratio &
smooth gate replaces hard clipping \\
GSPO~\citep{zheng2025gspo} & ratio granularity &
length-normalized sequence-level ratio \\
Entropy / cov.\ control~\citep{cui2025entropy} &
regularizer, gradient & bonus terms; detach or clip high-cov.\ tokens \\
CW-GRPO~\citep{wang2026cwgrpo} & token advantage &
Gaussian-kernel weight on log-prob/advantage covariance \\
\textbf{SCARL (ours)} & \textbf{advantage} &
\textbf{alignment-gap reshaping of the advantage} \\
\bottomrule
\end{tabular}
\end{table}

\paragraph{Confidence-aware update rules.}
Table~\ref{tab:update_rules} places SCARL next to the methods that
make the group-relative update depend on the policy's confidence in
each response (\S\ref{sec:related}), by the confidence signal used,
the quantity it acts on, the form of the modification, the outcome
classes it acts on, how the sign of the update is protected, and the
reference point against which confidence is measured.

\begin{table*}[t]
\centering
\caption{Confidence-aware modifications of the group-relative update.
``Sign'' states how the direction of the base update is protected;
``Reference'' states what each response's confidence is measured
against.}
\label{tab:update_rules}
\footnotesize
\setlength{\tabcolsep}{4pt}
\renewcommand{\arraystretch}{1.12}
\begin{tabular}{@{}>{\raggedright\arraybackslash}p{0.11\textwidth}
>{\raggedright\arraybackslash}p{0.16\textwidth}
>{\raggedright\arraybackslash}p{0.10\textwidth}
>{\raggedright\arraybackslash}p{0.20\textwidth}
>{\raggedright\arraybackslash}p{0.07\textwidth}
>{\raggedright\arraybackslash}p{0.10\textwidth}
>{\raggedright\arraybackslash}p{0.15\textwidth}@{}}
\toprule
Method & Signal & Acts on & Form & Classes & Sign & Reference \\
\midrule
CDE~\citep{dai2025cde} & response perplexity (length-normalized) &
reward & additive bonus, capped at $|r|/\kappa$, annealed & both &
by the cap & none (absolute) \\
ProGRPO / ARM~\citep{li2026progrpo} & low-probability-token,
length-normalized likelihood & advantage & additive
$\alpha\,(c(q) - c(o \mid q))$; off on all-wrong groups & both &
under a bound on $\alpha$ & group (prompt-level) confidence \\
CAPO~\citep{wang2026capo} & mean token log-prob & advantage
(replaced) & pairwise logistic-AUC surrogate; reference-model noise
masking & both & by construction & responses of the other class \\
ACE~\citep{xu2026ace} & $\log \pi_\theta(y \mid x) - \log
\pi_{\mathrm{ref}}(y \mid x)$ & negative advantages & scaled by a
softplus of the confidence shift & negatives & scaling & reference
model \\
CW-GRPO~\citep{wang2026cwgrpo} & token log-prob/advantage
covariance & token advantage & Gaussian-kernel weight, mean one
over the batch & both & scaling & batch covariance std \\
EAS~\citep{cheng2025reasoning} & token entropy & advantage &
additive, clipped, detached & both & by the clip & none \\
SEED-GRPO~\citep{chen2025seedgrpo} & semantic entropy of the
group's answers & advantage & multiplicative question-level factor
& both & scaling & none (per question) \\
UCAS~\citep{xie2025ucas} & logit-space self-certainty; token logits
& advantage & response-level modulation plus token-level penalty &
both & by a clamp & group statistics \\
\textbf{SCARL (ours)} & summed sequence log-likelihood,
group-standardized (Eq.~\eqref{eq:softmax}) & advantage &
multiplicative alignment gap $|t_i - p_i|$, normalized to mean one
per class (Eqs.~\eqref{eq:signaware}--\eqref{eq:classnorm}) & both
& for any $\tau$ (Prop.~\ref{prop:positivity}) & class-mean
confidence; exact recovery (Prop.~\ref{prop:recovery}) \\
\bottomrule
\end{tabular}
\end{table*}

\paragraph{Algorithm.}
Algorithm~\ref{alg:scarl} summarizes the SCARL procedure for one
prompt group; the reference implementation in
Appendix~\ref{app:impl} mirrors it line by line.

\begin{algorithm}[t]
\caption{SCARL advantage shaping (one prompt group)}
\label{alg:scarl}
\begin{algorithmic}[1]
\Require rollouts $\{(y_i, r_i)\}_{i=1}^{G}$ with stored per-token
log-probs under $\pi_{\theta_{\mathrm{old}}}$; base advantages
$\hat A_{i,t}$; hyperparameters $\tau, \delta$
\State $c_i \gets \sum_t \log
\pi_{\theta_{\mathrm{old}}}(y_{i,t} \mid x, y_{i,<t})$
\Comment{stored log-probs; no extra pass}
\State $\sigma_{\mathcal G} \gets \mathrm{std}(\{c_i\})$;\quad
$\tilde c_i \gets c_i / \big(\tau(\sigma_{\mathcal G} + \delta)\big)$
\State $p \gets \mathrm{softmax}(\tilde c)$
\Comment{subtract $\max_j \tilde c_j$ for stability}
\State $\tilde w_i \gets 1 - p_i$ if $\hat A_i \ge 0$
else $p_i$
\State $w_i \gets \tilde w_i \,/\, \mathrm{mean}_{j \in \mathcal S(i)}
\tilde w_j$
\Comment{per class; $w_i \gets 1$ if $|\mathcal S(i)| \le 1$}
\State \Return $\hat A^{\textsc{scarl}}_{i,t} = w_i \hat A_{i,t}$
for all $i, t$
\end{algorithmic}
\end{algorithm}

\FloatBarrier
% ---------------------------------------------------------------------
\section{Additional Results}
\label{supp:results}

\paragraph{Qwen3-1.7B-Base main comparison.}
Table~\ref{tab:main_pass32_17b} reports the full nine-method
comparison on Qwen3-1.7B-Base, complementing the Qwen3-4B-Base table
in the main paper. DAPO\,+\,SCARL attains the highest average
Pass@32 ($58.0$ vs.\ $56.4$ for the best baseline), with a same-base
margin of $+2.2$ over DAPO, wins AIME25 and Minerva outright, and
stays close to the best baseline on the remaining benchmarks.

\begin{table*}[t]
\centering
\caption{\textbf{Main comparison on Qwen3-1.7B-Base} (companion to
the main paper's Qwen3-4B-Base table). Pass@32 on five
competition-mathematics benchmarks (\%), at each method's best
checkpoint; \textbf{bold} marks the best value per column, with the
untrained base model excluded from bolding.}
\label{tab:main_pass32_17b}
\small
\setlength{\tabcolsep}{6pt}
\begin{tabular}{@{}lcccccc@{}}
\toprule
Method & AMC23 & AIME24 & AIME25 & Minerva & MATH-500 & Avg. \\
\midrule
Base model (no RL) & 71.2 & 14.7 & 11.4 & 48.0 & 87.8 & 46.6 \\
\addlinespace[2pt]
\multicolumn{7}{@{}l}{\itshape Base estimators}\\
GRPO & 79.1 & 31.9 & 22.6 & 49.3 & 89.5 & 54.5 \\
DAPO & \textbf{83.4} & 31.4 & 22.8 & 51.9 & 89.6 & 55.8 \\
GSPO & 83.3 & 28.7 & 27.2 & 52.8 & 90.0 & 56.4 \\
SAPO & 81.6 & 28.4 & 26.8 & 51.1 & 90.8 & 55.8 \\
Dr. GRPO & 79.3 & 30.6 & 26.2 & 50.8 & \textbf{90.9} & 55.6 \\
\addlinespace[2pt]
\multicolumn{7}{@{}l}{\itshape Confidence-aware methods}\\
Entropy Advantage Shaping & 78.2 & \textbf{35.3} & 22.9 & 50.6 & 90.2 & 55.4 \\
SEED-GRPO & 80.5 & 30.1 & 27.8 & 48.6 & 89.1 & 55.2 \\
UCAS & 73.3 & 18.2 & 10.8 & 50.2 & 87.9 & 48.1 \\
\addlinespace[2pt]
DAPO\,+\,SCARL (ours) & 82.7 & 34.7 & \textbf{29.4} & \textbf{53.2} & 89.8 & \textbf{58.0} \\
\bottomrule
\end{tabular}
\end{table*}

\paragraph{Per-benchmark generality results.}
Table~\ref{tab:generality_full} reports all sixteen
estimator--backbone pairings with per-benchmark scores; the main
paper's Table~\ref{tab:generality} summarizes the $\Delta$ column.

\begin{table*}[t]
\centering
\caption{\textbf{SCARL as a drop-in modifier of four base
estimators, across four backbones (full per-benchmark results).}
Pass@32 (\%) at each run's best checkpoint. Each estimator appears
as a base row and a \mbox{+\,SCARL} row; within every pair,
\textbf{bold} marks the higher per-column value, and $\Delta$ is the
change in average Pass@32
(\textcolor[HTML]{009E73}{gain}\,/\,\textcolor[HTML]{D55E00}{loss}),
computed from unrounded values. ``Base model (no RL)'' rows give
each untrained backbone as a reference.}
\label{tab:generality_full}
\small
\setlength{\tabcolsep}{4.5pt}
\begin{tabular}{@{}l cccccc c@{}}
\toprule
Method & AMC23 & AIME24 & AIME25 & Minerva & MATH-500 & Avg. & $\Delta$Avg. \\
\midrule
\multicolumn{8}{@{}l}{\bfseries Qwen3-4B-Base}\\
\addlinespace[1pt]
Base model (no RL) & 75.1 & 25.9 & 23.3 & 56.3 & 92.0 & 54.5 & \\
\addlinespace[1pt]
GRPO & 88.9 & 44.6 & \textbf{44.7} & 56.4 & 95.3 & 66.0 & \\
\quad + SCARL & \textbf{89.7} & \textbf{50.3} & 41.1 & \textbf{60.0} & \textbf{95.7} & \textbf{67.3} & \textcolor{deltaGain}{$\mathbf{+1.4}$} \\
DAPO & 90.4 & 46.1 & 45.0 & \textbf{58.4} & 95.5 & 67.1 & \\
\quad + SCARL & \textbf{95.3} & \textbf{49.0} & \textbf{50.1} & 57.4 & \textbf{95.6} & \textbf{69.5} & \textcolor{deltaGain}{$\mathbf{+2.4}$} \\
GSPO & 91.8 & 45.3 & \textbf{42.9} & 58.3 & 92.8 & 66.2 & \\
\quad + SCARL & \textbf{92.1} & \textbf{50.3} & 40.4 & \textbf{60.1} & \textbf{95.7} & \textbf{67.7} & \textcolor{deltaGain}{$\mathbf{+1.5}$} \\
SAPO & 92.2 & 45.7 & \textbf{43.1} & \textbf{59.3} & 94.8 & 67.0 & \\
\quad + SCARL & \textbf{92.3} & \textbf{53.4} & 41.0 & 58.3 & \textbf{95.6} & \textbf{68.1} & \textcolor{deltaGain}{$\mathbf{+1.1}$} \\
\midrule
\multicolumn{8}{@{}l}{\bfseries Qwen3-1.7B-Base}\\
\addlinespace[1pt]
Base model (no RL) & 71.2 & 14.7 & 11.4 & 48.0 & 87.8 & 46.6 & \\
\addlinespace[1pt]
GRPO & 79.1 & \textbf{31.9} & 22.6 & 49.3 & 89.5 & 54.5 & \\
\quad + SCARL & \textbf{81.2} & 29.2 & \textbf{28.3} & \textbf{52.1} & \textbf{90.6} & \textbf{56.3} & \textcolor{deltaGain}{$\mathbf{+1.8}$} \\
DAPO & \textbf{83.4} & 31.4 & 22.8 & 51.9 & 89.6 & 55.8 & \\
\quad + SCARL & 82.7 & \textbf{34.7} & \textbf{29.4} & \textbf{53.2} & \textbf{89.8} & \textbf{58.0} & \textcolor{deltaGain}{$\mathbf{+2.2}$} \\
GSPO & 83.3 & 28.7 & 27.2 & \textbf{52.8} & 90.0 & 56.4 & \\
\quad + SCARL & \textbf{84.8} & \textbf{28.7} & \textbf{29.0} & 52.3 & \textbf{90.6} & \textbf{57.1} & \textcolor{deltaGain}{$\mathbf{+0.7}$} \\
SAPO & \textbf{81.6} & \textbf{28.4} & 26.8 & 51.1 & 90.8 & \textbf{55.8} & \\
\quad + SCARL & 80.1 & 24.3 & \textbf{27.4} & \textbf{54.5} & \textbf{91.5} & 55.6 & \textcolor{deltaLoss}{$\mathbf{-0.2}$} \\
\midrule
\multicolumn{8}{@{}l}{\bfseries DeepSeek-R1-Distill-Qwen-1.5B}\\
\addlinespace[1pt]
Base model (no RL) & 68.8 & 26.1 & 28.0 & 52.2 & 93.2 & 53.7 & \\
\addlinespace[1pt]
GRPO & \textbf{95.4} & 59.5 & 40.6 & 54.9 & 96.0 & 69.3 & \\
\quad + SCARL & 92.2 & \textbf{62.7} & \textbf{46.4} & \textbf{55.3} & \textbf{96.1} & \textbf{70.5} & \textcolor{deltaGain}{$\mathbf{+1.2}$} \\
DAPO & 94.3 & 63.3 & 47.9 & 54.9 & \textbf{96.8} & 71.4 & \\
\quad + SCARL & \textbf{95.0} & \textbf{64.0} & \textbf{48.6} & \textbf{56.3} & 96.0 & \textbf{72.0} & \textcolor{deltaGain}{$\mathbf{+0.5}$} \\
GSPO & \textbf{95.6} & 64.7 & 42.6 & \textbf{55.3} & \textbf{96.3} & 70.9 & \\
\quad + SCARL & 93.8 & \textbf{67.4} & \textbf{45.9} & 54.3 & 95.4 & \textbf{71.4} & \textcolor{deltaGain}{$\mathbf{+0.5}$} \\
SAPO & 93.6 & 62.0 & \textbf{50.8} & 55.0 & \textbf{96.2} & 71.5 & \\
\quad + SCARL & \textbf{94.7} & \textbf{67.1} & 45.5 & \textbf{55.1} & 95.8 & \textbf{71.6} & \textcolor{deltaGain}{$\mathbf{+0.1}$} \\
\midrule
\multicolumn{8}{@{}l}{\bfseries Hunyuan-1.8B-Pretrain}\\
\addlinespace[1pt]
Base model (no RL) & 82.6 & 28.2 & 27.8 & 47.5 & 92.2 & 55.6 & \\
\addlinespace[1pt]
GRPO & 85.5 & \textbf{42.2} & 33.6 & 50.8 & 94.0 & 61.2 & \\
\quad + SCARL & \textbf{88.3} & 38.9 & \textbf{35.6} & \textbf{52.8} & \textbf{94.3} & \textbf{62.0} & \textcolor{deltaGain}{$\mathbf{+0.8}$} \\
DAPO & 85.8 & 44.0 & \textbf{38.5} & 52.4 & 94.0 & 62.9 & \\
\quad + SCARL & \textbf{89.5} & \textbf{49.1} & 35.8 & \textbf{52.9} & \textbf{94.8} & \textbf{64.4} & \textcolor{deltaGain}{$\mathbf{+1.5}$} \\
GSPO & 86.1 & \textbf{45.0} & 37.4 & 48.8 & 92.5 & 62.0 & \\
\quad + SCARL & \textbf{86.3} & 39.7 & \textbf{37.6} & \textbf{53.0} & \textbf{94.0} & \textbf{62.1} & \textcolor{deltaGain}{$\mathbf{+0.1}$} \\
SAPO & 83.9 & 41.4 & \textbf{42.9} & 52.0 & 93.3 & \textbf{62.7} & \\
\quad + SCARL & \textbf{87.8} & \textbf{43.3} & 33.6 & \textbf{53.3} & \textbf{94.5} & 62.5 & \textcolor{deltaLoss}{$\mathbf{-0.2}$} \\
\bottomrule
\end{tabular}
\end{table*}

\paragraph{Where the gains concentrate.}
Averaged over estimators, the improvement is largest on the two
Qwen3 backbones ($+1.6$ on the 4B model, $+1.1$ on the 1.7B model)
and smaller though still positive on DeepSeek-R1-Distill-Qwen-1.5B
and Hunyuan-1.8B-Pretrain ($+0.6$ each). Averaged over backbones, it
is largest on DAPO ($+1.7$) and GRPO ($+1.3$) and smaller on GSPO
($+0.7$) and SAPO ($+0.2$). The model-level pattern is consistent
with headroom in the quantity SCARL acts on: the distilled backbone
inherits its teacher's reasoning distribution and already reaches
the highest coverage of the four models ($69.3$--$71.5$ average
Pass@32 before SCARL), and the Hunyuan backbone starts from an
already-high confidence--correctness separation; both leave less
misalignment for a confidence-aware reweighting to repair. The
base-model rows put the same picture on an absolute scale: every one
of the $32$ trained runs raises coverage over its untrained backbone
($+5.6$ to $+18.3$ points), and adding SCARL widens that margin in
$14$ of the $16$ pairings.

\paragraph{Per-benchmark ablation results.}
Table~\ref{tab:ablation_full} gives the per-benchmark breakdown of
the component ablations and temperature sweep summarized in the main
paper.

\begin{table*}[t]
\centering
\caption{\textbf{Component ablations and temperature sensitivity,
full per-benchmark results} (Qwen3-1.7B-Base, GSPO base estimator).
Pass@32 (\%) at each run's best checkpoint; $\Delta$ is the change
in average Pass@32 relative to plain GSPO, computed from unrounded
values. The default $\tau = 5$ row duplicates the full-SCARL run for
reference.}
\label{tab:ablation_full}
\small
\setlength{\tabcolsep}{4.5pt}
\begin{tabular}{@{}l cccccc c@{}}
\toprule
Variant & AMC23 & AIME24 & AIME25 & Minerva & MATH-500 & Avg. & $\Delta$Avg. \\
\midrule
\multicolumn{8}{@{}l}{\itshape Component ablations}\\
\addlinespace[1pt]
GSPO (base estimator) & 83.3 & 28.7 & 27.2 & 52.8 & 90.0 & 56.4 & --- \\
\quad + SCARL w/o class-wise norm. & 65.8 & 15.8 & 16.8 & 49.0 & 87.6 & 47.0 & \textcolor{deltaLoss}{$\mathbf{-9.4}$} \\
\quad + SCARL w/o alignment target & 83.1 & 34.5 & 22.6 & 52.0 & 90.6 & 56.6 & \textcolor{deltaGain}{$\mathbf{+0.2}$} \\
\quad + SCARL w/ per-token mean confidence & 79.9 & 32.7 & 20.9 & 52.7 & 90.1 & 55.3 & \textcolor{deltaLoss}{$\mathbf{-1.1}$} \\
\quad + SCARL (full) & 84.8 & 28.7 & 29.0 & 52.3 & 90.6 & 57.1 & \textcolor{deltaGain}{$\mathbf{+0.7}$} \\
\midrule
\multicolumn{8}{@{}l}{\itshape Temperature sweep (full SCARL)}\\
\addlinespace[1pt]
\quad $\tau = 0.2$ & 78.1 & 30.3 & 25.5 & 50.7 & 90.8 & 55.1 & \textcolor{deltaLoss}{$\mathbf{-1.3}$} \\
\quad $\tau = 0.5$ & 77.9 & 28.0 & 29.2 & 53.7 & 89.0 & 55.6 & \textcolor{deltaLoss}{$\mathbf{-0.8}$} \\
\quad $\tau = 2$ & 80.8 & 33.7 & 23.8 & 52.4 & 90.5 & 56.2 & \textcolor{deltaLoss}{$\mathbf{-0.2}$} \\
\quad $\tau = 5$ (default) & 84.8 & 28.7 & 29.0 & 52.3 & 90.6 & 57.1 & \textcolor{deltaGain}{$\mathbf{+0.7}$} \\
\quad $\tau = 10$ & 75.4 & 28.0 & 27.6 & 52.5 & 90.1 & 54.7 & \textcolor{deltaLoss}{$\mathbf{-1.7}$} \\
\bottomrule
\end{tabular}
\end{table*}

\paragraph{Mechanism, per-step statistics.}
For the settings of Figure~\ref{fig:mechanism}, SCARL's Cohen's $d$
exceeds the base run's at $83\%$ (Qwen3-4B-Base\,/\,DAPO), $95\%$
(DeepSeek-R1-Distill-Qwen-1.5B\,/\,GSPO), and $92\%$
(Hunyuan-1.8B-Pretrain\,/\,GRPO) of the $150$ steps, and over the
final fifty steps at $149$ of the $150$ setting-steps. The one
systematic deviation is the early phase of the
Qwen3-4B-Base\,/\,DAPO setting, where the base run's $d$ is
transiently higher for much of the first $35$ steps before SCARL
overtakes it at $96\%$ of the remaining steps; this is consistent
with the design, which damps precisely the confident-success
reinforcement that inflates separation fastest at the start. On the
entropy side, the final-ten-step means (base vs.\ +SCARL) are
$0.035$/$0.050$ on GRPO, $0.50$/$1.02$ on DAPO, $0.12$/$0.49$ on
GSPO, and $0.07$/$0.73$ on SAPO.
\FloatBarrier

% scarl_appendix.tex: technical appendix (input by supplement.tex,
% which issues \appendix). Listings use the kit's ruled listing float.

% ---------------------------------------------------------------------
\section{Formal Statements and Proofs}
\label{app:proofs}

This appendix states and proves the guarantees asserted in
\S\ref{sec:classnorm}. Throughout, fix one prompt group, let
$p \in (0,1)^G$ be the softmax output of Eq.~\eqref{eq:softmax}, let
$\tilde w_i$ be the alignment gap of Eq.~\eqref{eq:signaware}, let
$\mathcal S(i) \in \{\mathcal S_{+}, \mathcal S_{-}\}$ be the target
class of sample $i$, and let $w_i$ be the normalized shaping factor of
Eq.~\eqref{eq:classnorm}, with $w_i = 1$ whenever
$|\mathcal S(i)| \le 1$.

\begin{proposition}[Positivity and boundedness; Property~1]
\label{prop:positivity}
For every $i$, $w_i > 0$; consequently
$\operatorname{sign}\big(w_i \hat A_{i,t}\big) =
\operatorname{sign}\big(\hat A_{i,t}\big)$ for every token $t$.
Moreover, if $|\mathcal S(i)| \ge 2$ then $w_i < |\mathcal S(i)|$.
\end{proposition}

\begin{proof}
Softmax outputs of finite logits satisfy $p_i \in (0,1)$ strictly, so
$\tilde w_i \in \{1 - p_i,\, p_i\} \subset (0,1)$ and every raw
gap is strictly positive. If $|\mathcal S(i)| \le 1$, then
$w_i = 1 > 0$. Otherwise the class mean
$\frac{1}{|\mathcal S(i)|} \sum_{j \in \mathcal S(i)} \tilde w_j$ is
a mean of strictly positive numbers, hence strictly positive, and
$w_i > 0$ follows from Eq.~\eqref{eq:classnorm}. Multiplying
$\hat A_{i,t}$ by a strictly positive scalar preserves its sign (and
maps zero to zero). For the bound, rewrite
Eq.~\eqref{eq:classnorm} as
$w_i = |\mathcal S(i)|\, \tilde w_i \big/
\sum_{j \in \mathcal S(i)} \tilde w_j$; since the sum contains, in
addition to $\tilde w_i$, at least one other strictly positive term,
$\tilde w_i < \sum_{j \in \mathcal S(i)} \tilde w_j$ and therefore
$w_i < |\mathcal S(i)|$.
\end{proof}

\begin{proposition}[Class-mass preservation; Property~2, first part]
\label{prop:classmass}
For each nonempty class $\mathcal S$,
$\frac{1}{|\mathcal S|} \sum_{i \in \mathcal S} w_i = 1$.
\end{proposition}

\begin{proof}
If $|\mathcal S| = 1$ this holds by the convention $w_i = 1$ (and is
also what Eq.~\eqref{eq:classnorm} returns, since the class mean then
equals $\tilde w_i$ itself). If $|\mathcal S| \ge 2$, summing
Eq.~\eqref{eq:classnorm} over $i \in \mathcal S$ gives
$\sum_{i \in \mathcal S} w_i =
\big(\sum_{i \in \mathcal S} \tilde w_i\big) \big/
\big(\frac{1}{|\mathcal S|} \sum_{j \in \mathcal S} \tilde w_j\big)
= |\mathcal S|$. This is a deterministic identity; the notation
$\mathbb{E}[\, w_i \mid \mathcal S(i)\,] = 1$ in
\S\ref{sec:classnorm} denotes exactly this within-class empirical
mean.
\end{proof}

\begin{proposition}[Exact recovery; Property~2, second part]
\label{prop:recovery}
If the confidences $c_i$ are constant within each target class, then
$w_i = 1$ for every sample, and the SCARL objective coincides exactly
with the base estimator's. In particular this holds when all $c_i$ in
the group are equal.
\end{proposition}

\begin{proof}
The map from $c_i$ to the scaled logit $\tilde c_i$ in
Eq.~\eqref{eq:softmax} applies the same strictly monotone affine
transformation to every group member, so equal confidences yield
equal logits, and the softmax assigns equal probabilities to equal
logits. Hence $p_i$, and with it $\tilde w_i$, is constant within
each target class. Eq.~\eqref{eq:classnorm} then divides each
$\tilde w_i$ by the class mean of identical values, giving
$w_i = 1$; classes of size at most one give $w_i = 1$ by convention.
With all factors equal to one,
$\hat A^{\textsc{scarl}}_{i,t} = \hat A_{i,t}$, so
$\mathcal{J}_{\mathrm{base}}\big(\theta; \{w_i \hat A_{i,t}\}\big) =
\mathcal{J}_{\mathrm{base}}\big(\theta; \{\hat A_{i,t}\}\big)$. When
all $c_i$ in the group are equal we have
$\sigma_{\mathcal G} = 0$ and every scaled logit equals
$c_i / (\tau \delta)$, a single shared value, so $p$ is uniform
and the same argument applies.
\end{proof}

\noindent\emph{Remark.}
Shaping is active only where confidence varies within a target
class, which is precisely the dispersion the alignment criterion of
\S\ref{sec:method} concerns.

\begin{remark}[Scale imbalance without normalization]
\label{rem:imbalance}
This makes precise the calculation asserted in
\S\ref{sec:classnorm}. If $p$ is exactly uniform, $p_i = 1/G$, the
raw gaps of Eq.~\eqref{eq:signaware} are $(G-1)/G$ for every
positive sample and $1/G$ for every negative one: a ratio of
$(G-1)$ to $1$ between the classes, i.e., $7\times$ at our group
size $G = 8$. Used without Eq.~\eqref{eq:classnorm}, the raw gaps
would therefore rescale the two classes' advantage budgets by these
systematically different factors, all but erasing penalties; the
class-wise normalization removes exactly this distortion while
leaving the within-class allocation untouched. The ablation of
\S\ref{sec:ablation} measures the consequence of omitting it
($-9.4$ average Pass@32, below the untrained backbone).
\end{remark}

\FloatBarrier
% ---------------------------------------------------------------------
\section{Reference Implementation and Practical Notes}
\label{app:impl}

Algorithm~\ref{alg:scarl} maps to a short, framework-agnostic
function operating on one prompt group.
Listing~\ref{lst:impl} mirrors it line by line in PyTorch style; it
consumes quantities every group-relative trainer already
materializes and touches nothing else.

\begin{listing}[tb]
\caption{Reference implementation of SCARL advantage shaping
(PyTorch). The released code names the stabilizer \texttt{eps},
written $\delta$ in Eq.~\eqref{eq:softmax}.}
\label{lst:impl}
\begin{lstlisting}[language=Python,numbers=left,xleftmargin=1.5em,numbersep=6pt,breaklines=false]
def scarl_advantages(adv, old_lp, mask,
                     tau=5.0, eps=1e-8):
    # adv:    (G,)  base seq. advantages
    # old_lp: (G,T) behavior-policy token
    #   log-probs (same source as the
    #   importance ratios)
    # mask:   (G,T) response-token mask
    c = (old_lp * mask).sum(-1)  # eq:conf
    sd = c.std(correction=0)
    z = c / (tau * (sd + eps))
    p = torch.softmax(z, 0)      # eq:softmax
    pos = adv >= 0
    w_raw = torch.where(pos, 1 - p, p)
    w = torch.ones_like(w_raw)
    for cls in (pos, ~pos):      # class-norm
        if cls.sum() >= 2:
            w[cls] = w_raw[cls] \
                     / w_raw[cls].mean()
    return w * adv  # eq:scarl; broadcast
                    # to tokens downstream
\end{lstlisting}
\end{listing}

\paragraph{Practical notes.}
(i)~\emph{Tie-breaking.} Samples with $\hat A_i = 0$ are assigned to
$\mathcal S_{+}$ by convention. Groups whose rewards are all equal
yield $\hat A_i = 0$ for every sample under GRPO, so the shaping is
inert there regardless of the factors; such groups are in any case
filtered out under DAPO-style dynamic sampling.
(ii)~\emph{Log-prob source.} If the trainer recomputes old-policy
log-probabilities (for example, under an inference-engine and
trainer mismatch), compute $c_i$ from the same log-probabilities
used for the importance ratios, so that the shaping factor and the ratio
describe the same behavior policy.
(iii)~\emph{Epoch constancy.} Because $c_i$ depends only on
$\pi_{\theta_{\mathrm{old}}}$, the shaping factors are computed once per
rollout and reused unchanged across all minibatch updates of the
epoch; no gradient flows through them.
(iv)~\emph{Standard-deviation convention.} Population and sample
standard deviations differ only by a factor shared across the group,
which is absorbed into the effective temperature
$\tau(\sigma_{\mathcal G} + \delta)$; either convention works if
used consistently.
(v)~\emph{Stabilizer form.} The released implementation guards the
scaling with $\max(\sigma_{\mathcal G}, 10^{-8})$ rather than the
additive $\sigma_{\mathcal G} + \delta$ of
Eq.~\eqref{eq:softmax}; the two coincide except on degenerate groups
with near-zero confidence dispersion, where both reduce SCARL toward
uniform factors.
(vi)~\emph{Configuration naming.} The released scripts and trainer
expose alignment-gap shaping (\S\ref{sec:signaware}) under the
historical flag \texttt{scarl\_sign\_aware\_weighting}; it is the
same computation, with the case split of
Eq.~\eqref{eq:signaware} implemented branchwise.

\FloatBarrier
% ---------------------------------------------------------------------
\section{Training and Evaluation Details}
\label{app:hparams}

Table~\ref{tab:hparams} lists the complete shared configuration,
including the values quoted in \S\ref{sec:setup}; all methods
use it identically, differing only in the surrogate (base estimators,
following their published recipes, including each recipe's KL choice)
or in the advantage transformation (SCARL).

\begin{table}[htbp]
\centering
\caption{Shared training and evaluation configuration. Every method
in every experiment uses these values; per-estimator settings appear
in Table~\ref{tab:estimator_hparams}.}
\label{tab:hparams}
\footnotesize
\setlength{\tabcolsep}{4pt}
\begin{tabular}{@{}l>{\raggedright\arraybackslash}p{0.51\columnwidth}@{}}
\toprule
\multicolumn{2}{@{}l}{\itshape Training} \\
Framework & verl v0.4.1~\citep{sheng2024hybridflow} \\
Prompt set & DAPO-Math-17K~\citep{yu2025dapo} \\
Reward & $+1$ verified correct, $-1$ otherwise \\
Verifier & rule-based, per data source
(Appendix~\ref{app:prompts}) \\
Training steps & $150$ \\
Group size $G$ & $8$ \\
Prompt batch size & $512$ \\
Mini-batch size & $32$ prompts ($16$ updates per step) \\
Optimizer & AdamW, weight decay $0.1$ \\
Learning rate & $1 \times 10^{-6}$, constant after
$10$ warmup steps \\
Gradient clipping & $1.0$; entropy coefficient $0$ \\
Max prompt length & $2{,}048$ tokens \\
Max response length & $4{,}096$ tokens \\
Rollout sampling (training) & temperature $1.0$, top-$p$ $1.0$ \\
Rollout engine & vLLM; FSDP training backend \\
\addlinespace[2pt]
\multicolumn{2}{@{}l}{\itshape SCARL} \\
Temperature $\tau$ & $5$ (sweep in \S\ref{sec:ablation});
adaptive scaling on \\
Confidence source & old-policy token log-probs
(Eq.~\eqref{eq:conf}) \\
Stabilizer & $\sigma_{\mathcal G}$ floored at $10^{-8}$
(Appendix~\ref{app:impl}) \\
\addlinespace[2pt]
\multicolumn{2}{@{}l}{\itshape Evaluation} \\
Sampling & temperature $0.6$, top-$p$ $0.95$, top-$k$ $20$ \\
Max generation length & $4{,}096$ response tokens \\
Responses per problem $k$ & $32$ \\
Evaluation cadence & every $5$ steps ($30$ checkpoints) \\
Checkpoint selection & best five-benchmark average \\
Pass@32 estimator & bootstrap over the $32$ responses
($1{,}000$ resamples) \\
\bottomrule
\end{tabular}
\end{table}

\begin{table}[htbp]
\centering
\caption{Per-estimator surrogate settings, following each method's
published recipe (extracted from the released run scripts). All rows
share the configuration of Table~\ref{tab:hparams}.}
\label{tab:estimator_hparams}
\footnotesize
\setlength{\tabcolsep}{4pt}
\begin{tabular}{@{}lp{0.62\columnwidth}@{}}
\toprule
Estimator & Settings \\
\midrule
GRPO & symmetric clip $\varepsilon = 0.2$; KL loss toward the
reference policy, coefficient $0.001$ (low-variance estimator);
token-mean aggregation \\
DAPO & $\varepsilon_{\mathrm{low}} = 0.2$,
$\varepsilon_{\mathrm{high}} = 0.28$; dynamic sampling (filter by
accuracy, generation batch $2{,}048$, at most $10$ generation
rounds); soft overlong punishment (buffer $4{,}096$ tokens, penalty
factor $1.0$); token-mean aggregation; no KL \\
GSPO & sequence-level ratio with clip range
$3 \times 10^{-4}$ / $4 \times 10^{-4}$; no KL \\
SAPO & smooth gate with temperatures
$\tau_{\mathrm{pos}} = 1.0$, $\tau_{\mathrm{neg}} = 1.05$ (published
defaults); no hard clipping; no KL \\
Dr.~GRPO & advantage without std normalization;
global-constant token-loss aggregation
(seq-mean-token-sum-norm); no KL \\
\bottomrule
\end{tabular}
\end{table}

\paragraph{Confidence-aware baselines.}
The three baselines run in their native configurations on their own
bases, with the shared recipe of Table~\ref{tab:hparams}. Entropy
Advantage Shaping~\citep{cheng2025reasoning} uses the paper's GRPO
setting, $\alpha = 0.4$ and $\kappa = 2.0$, adding the clipped,
detached entropy term to a standard GRPO advantage with vanilla clip
loss. SEED-GRPO~\citep{chen2025seedgrpo} modulates a Dr.~GRPO base
advantage with linear semantic-entropy weighting,
$f(\tilde s) = 1 - \alpha \tilde s$ with
$\tilde s = \mathrm{SE}/\log K$ and $\alpha = 0.02$ (the paper's
best), clustering responses by extracted final answer.
For UCAS~\citep{xie2025ucas}, we ran both the official repository's
default configuration and the paper's recommended settings, and
report the better performing of the two, the repository defaults.

\paragraph{Compute.}
Every run uses a single node with $8$ NVIDIA A100 40GB GPUs, with
vLLM rollout and FSDP training (parameter and optimizer offload,
gradient checkpointing, sequence parallel size $1$).

\paragraph{Pass@32 estimator.}
Pass@32 is computed with verl's bootstrap
estimator~\citep{sheng2024hybridflow}: for each problem, the
any-correct indicator is averaged over $1{,}000$ size-$32$ resamples
drawn with replacement from the problem's $32$ responses, which
equals $1 - (1 - c/32)^{32}$ in expectation for $c$ correct
responses; the per-problem values are then averaged over the
benchmark.

% ---------------------------------------------------------------------
\subsection{Prompt Templates}
\label{app:prompts}

Each prompt is presented as a single user turn through the
tokenizer's chat template. Two conventions are in use, matching
each benchmark's rule-based scorer;
Listings~\ref{lst:prompt_answerline} and~\ref{lst:prompt_boxed}
give the templates, with \texttt{\{problem\}} marking the inserted
problem statement.

\paragraph{Answer-line convention.}
The training set (DAPO-Math-17K~\citep{yu2025dapo}, data source
\texttt{math\_dapo}) and
the AIME24, AIME25, and AMC23 sets use the template of
Listing~\ref{lst:prompt_answerline}, instructing the model to end
with an answer line; correctness is checked by extracting the final
\texttt{Answer:} line (DAPO's rule-based scorer, with the answer
normalization of \texttt{math\_dapo.py}; AMC23 is routed to this
same scorer).

\begin{listing}[tb]
\caption{Answer-line prompt template (training set, AIME24, AIME25,
and AMC23); a single paragraph in the data, wrapped here for
display.}
\label{lst:prompt_answerline}
\begin{lstlisting}
Solve the following math problem step by step. The last line of your response should be of the form Answer: $Answer (without quotes) where $Answer is the answer to the problem.

{problem}

Remember to put your answer on its own line after "Answer:".
\end{lstlisting}
\end{listing}

\paragraph{Boxed convention.}
The MATH-500 and Minerva Math sets append the boxed instruction of
Listing~\ref{lst:prompt_boxed} to the problem; correctness is
checked with
Math-Verify.\footnote{\url{https://github.com/huggingface/Math-Verify}}

\begin{listing}[tb]
\caption{Boxed prompt template (MATH-500 and Minerva Math).}
\label{lst:prompt_boxed}
\begin{lstlisting}
{problem}
Let's think step by step and output the final answer within \boxed{}.
\end{lstlisting}
\end{listing}

\FloatBarrier
% ---------------------------------------------------------------------
\section{Benchmark Details}
\label{app:benchmarks}

Table~\ref{tab:benchmarks} lists the evaluation sets. MATH-500 is
the $500$-problem evaluation subset selected by
\citet{lightman2024letsverify} from the MATH
dataset~\citep{hendrycks2021math}; Minerva Math is the
quantitative-reasoning evaluation set of
\citet{lewkowycz2022minerva}; AMC23, AIME24, and AIME25 collect the
public competition problems of the corresponding years.

\begin{table}[htbp]
\centering
\caption{Evaluation benchmarks. Pass@32 and Avg@32 are averaged over
each benchmark's problems (\S\ref{sec:setup}).}
\label{tab:benchmarks}
\small
\setlength{\tabcolsep}{5pt}
\begin{tabular}{@{}lr@{}}
\toprule
Benchmark & Problems \\
\midrule
AMC23 & 40 \\
AIME24 & 30 \\
AIME25 & 30 \\
Minerva Math & 272 \\
MATH-500 & 500 \\
\bottomrule
\end{tabular}
\end{table}

\end{document}